\documentclass[10pt,twocolumn,letterpaper]{article}

\usepackage{cvpr}
\usepackage{times}
\usepackage{epsfig}
\usepackage{graphicx}
\usepackage{amsmath}
\usepackage{amssymb}
\usepackage[noend]{algpseudocode}
\usepackage{algorithmicx,algorithm}
\usepackage{multirow}
\usepackage{booktabs}

\usepackage[pagebackref=true,breaklinks=true,letterpaper=true,colorlinks,bookmarks=false]{hyperref}

\cvprfinalcopy

\def\cvprPaperID{xxxx} 

\begin{document}

\title{Learn to Augment: Joint Data Augmentation and Network Optimization \\ for Text Recognition}

\author{Canjie Luo\textsuperscript{\rm 1}, Yuanzhi Zhu\textsuperscript{\rm 1}, Lianwen Jin\textsuperscript{\rm 1}\thanks{Corresponding author.}{ \ },  Yongpan Wang\textsuperscript{\rm 2}\\
\textsuperscript{\rm 1}South China University of Technology, \textsuperscript{\rm 2}Alibaba Group\\
{\tt\small \{canjie.luo, zzz.yuanzhi, lianwen.jin\}@gmail.com, yongpan@taobao.com}}

\maketitle

\begin{abstract}
Handwritten text and scene text suffer from various shapes and distorted patterns. Thus training a robust recognition model requires a large amount of data to cover diversity as much as possible. In contrast to data collection and annotation, data augmentation is a low cost way. In this paper, we propose a new method for text image augmentation. Different from traditional augmentation methods such as  rotation, scaling and perspective transformation, our proposed augmentation method is designed to learn proper and efficient data augmentation which is more effective and specific for training a robust recognizer. By using a set of custom fiducial points, the proposed augmentation method is flexible and controllable. Furthermore, we bridge the gap between the isolated processes of data augmentation and network optimization by joint learning. An agent network learns from the output of the recognition network and controls the fiducial points to generate more proper training samples for the recognition network. Extensive experiments on various benchmarks, including regular scene text, irregular scene text and handwritten text, show that the proposed augmentation and the joint learning methods significantly boost the performance of the recognition networks. A general toolkit for geometric augmentation is available\footnote{https://github.com/Canjie-Luo/Text-Image-Augmentation}.
\end{abstract}

\section{Introduction}

The last decade witnessed the tremendous progress brought by the deep neural network in the computer vision community \cite{bahdanau2014neural,goodfellow2014generative,he2016deep,Krizhevsky2012ImageNet}. Limited data is not sufficient to train a robust deep neural network, because the network may overfit to the training data and produce poor generalization on the test set \cite{bhunia2019handwriting}. However, data collection and annotation require a lot of resources. Different from single object classification task \cite{Krizhevsky2012ImageNet}, the annotation work of text string is more tough, because there are multiple characters in a text image. This is also a reason why most state-of-the-art scene text recognition methods \cite{liao2019scene,cluo2019moran,shi2018aster} only used synthetic samples \cite{gupta2016synthetic,Jaderberg2015Reading} for training. The data limitation also effects handwritten text recognition. There exists a wide variety of writing styles. Collecting large scale annotated handwritten text image is high-cost and cannot cover all diversities \cite{zhang2019sequence}. It is also challenging to generate synthetic data for handwritten text, because it is difficult to imitate various writing styles.

\begin{figure}[t]
\centering
\includegraphics[width=1.0\columnwidth]{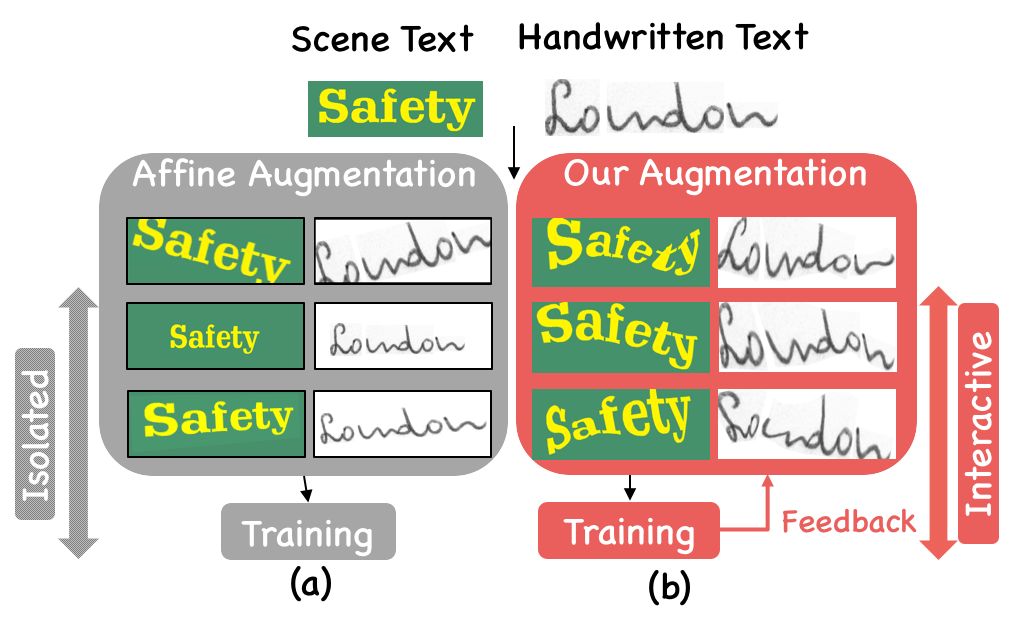} 
\caption{(a) Existing geometric augmentation, including rotation, scaling and perspective transformation; (b) Our proposed flexible augmentation. Moreover, a joint learning method bridges the isolated processes of data augmentation and network training.}
\label{pic-introduction}
\end{figure}

To obtain more training samples, it is possible to apply random augmentation to the existing data \cite{Cubuk2019AutoAugment}. Handwritten text with varying writing styles, and scene text with different shapes, such as perspective and curved text, are still very challenging to be recognized \cite{bhunia2019handwriting,cluo2019moran,shi2018aster}. Therefore, geometric augmentation is an important way to gain robustness for recognition methods. As shown in Figure \ref{pic-introduction} (a),  the common geometric transformations are rotation, scaling and perspective transformation. Multiple characters in an image are regarded as one entity, and a global augmentation is performed on the image. However, the diversity of each character should be taken into account. Given a text image, the augmentation goal is to increase the diversity of every character in the text string. Therefore, existing augmentation is limited to the over-simple transformations, which are inefficient for training.

In addition, the effective training samples that contribute to the robustness of the network may still be rare because of the long-tail distribution \cite{Peng2018Jointly}, which is another reason that causes inefficient training. The strategy of random augmentation is the same for every training sample, neglecting the difference among the samples and the optimization procedure of the network. Under the manually controlled static distribution, the augmentation may produce many ``easy" samples which are useless for the training. Therefore, random augmentation under the static distribution can hardly meet the requirement of the dynamic optimization. Simultaneously, the manually designed best augmentation strategy on a dataset, usually cannot be transferred to another dataset as expected. Our goal is to study a learnable augmentation method that can automatically adapt to other tasks without any manual modification.

In this paper, we propose a new data augmentation method for text recognition, which is designed for sequence-like characters \cite{shi2017end} augmentation. Our augmentation method focuses on the spatial transformation of images. We first initialize a set of fiducial points on the image and then move the points to generate a new image. The moving state, which represents the movement of the points to create ``harder" training samples, is sampled from the predicted distribution of the agent network. Then the augmentation module takes the moving state and image as input, and generates a new image. We adopt similarity transformation based on moving least squares \cite{Schaefer2006Image} for image generation. Besides, a random moving state is also fed to the augmentation module to generate a randomly augmented image. Finally, the agent learns from the moving state that increases recognition difficulty. The difficulty is measured under the metric of edit distance, which is highly relevant to the recognition performance.

To summarize, our contributions are as follows:
\begin{itemize}
\setlength{\itemsep}{0.5pt}
\setlength{\parsep}{1pt}
\setlength{\parskip}{1pt}
\item We propose a data augmentation method for text images that contain multiple characters. To the best of our knowledge, this may be the first augmentation method specially designed for sequence-like characters. 
\item We propose a framework that jointly optimizes the data augmentation and the recognition model. The augmented samples are generated through an automatic learning process, and are thus more effective and useful for the model training. The proposed framework is end-to-end trainable without any fine-tuning.
\item Extensive experiments conducted on various benchmarks, including scene text and handwritten text, show that the proposed augmentation and joint learning methods remarkably boost the performance of the recognizers, especially on small training dataset.
\end{itemize}

\section{Related Work}

\textbf{Scene Text Recognition} 
As an essential process in computer vision tasks, scene text recognition has attracted much research interest \cite{li2018show,liao2019scene,cluo2019moran,shi2018aster}. There are multiple characters in a scene text image. Thus the text string recognition task is more difficult than single character recognition. Typically, scene text recognition approaches can be divided into two types: localization-based and segmentation-free. 

The former attempts to localize the position of characters, recognize them and group all the characters as a text string \cite{wang2011end,wang2012end}. The latter benefits from the success of deep neural network and models the text recognition as a sequence recognition problem. For instance, He et al. \cite{he2016reading} and Shi et al. \cite{shi2017end} applied recurrent neural networks (RNNs) on the top of convolutional neural networks (CNNs) for spatial dependencies of sequence-like objects. Furthermore, the sequence-to-sequence mapping issue was addressed by attention mechanism \cite{shi2018aster}.

The great progress in regular text recognition led the community to irregular text recognition. Luo et al. \cite{cluo2019moran} and Shi et al. \cite{shi2018aster} proposed rectification networks to remove distortion and decrease recognition difficulty. Zhan and Lu \cite{Zhan2018ESIR} iteratively removed perspective distortion and text line curvature. Yang et al. \cite{yang2019symmetry} gave an accurate description of text shape by using more geometric constraints and supervisions for every character. Though the methods above made a notable step forward, irregular scene text recognition still remains a challenging problem.

\textbf{Handwritten Text Recognition} 
Due to various writing styles, handwritten text recognition is still a challenging field \cite{bhunia2019handwriting}. Early methods used hybrid hidden Markov model \cite{espana2010improving} and embedded both word images and text strings in a common vectorial subspace to cast recognition tasks as nearest neighbor problems \cite{almazan2014word}. 

In the deep learning era, Sueiras et al. \cite{sueiras2018offline} and Sun et al. \cite{sun2016convolutional} extracted feature by using CNNs followed by RNNs, and obtained superior results. Zhang et al. \cite{zhang2019sequence} addressed handwriting style diversity problem by proposing a sequence-to-sequence domain adaptation Network. Bhunia et al. \cite{bhunia2019handwriting} adversarially warped the intermediate feature-space to alleviate the lack of variations in some sparse training datasets. While great progress has been made, handwritten text recognition remains an open and challenging problem because of various writing styles. 

\begin{figure*}[t]
\centering
\includegraphics[width=2\columnwidth]{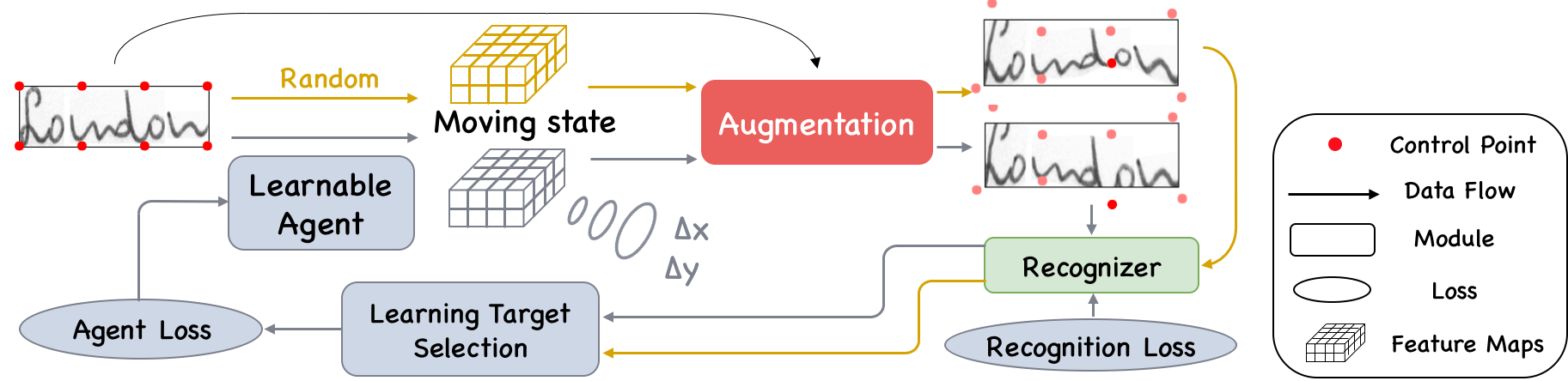} 
\caption{Overview of the proposed framework. First, the learnable agent predicts a distribution of the moving state aiming to create a harder training sample. Then the augmentation module generates augmented samples based on the random and predicted moving state, respectively. The difficulty of the pair of samples is measured by the recognition network. Finally, the agent takes the moving state that increases difficulty as guidance and updates itself. The unified framework is end-to-end trainable.}
\label{pic-overall}
\end{figure*}

\textbf{Data Augmentation} 
Data augmentation is critical to avoid overfitting in the training of deep neural networks \cite{Cubuk2019AutoAugment,Ho2019Population,Peng2018Jointly}. Nevertheless, few research addresses the augmentation issue for text images. Common geometric augmentations including flipping, rotation, scaling and perspective transformation, are typically useful for single object recognition \cite{Krizhevsky2012ImageNet}. However, a text image contains multiple characters. Existing over-simple transformations do not significantly contribute to the diversity of text appearance.

Simultaneously, the static augmentation policy does not meet the dynamic requirement of optimization. Cubuk et al. \cite{Cubuk2019AutoAugment} searched the policy for augmentation by using reinforcement learning. Ho et al. \cite{Ho2019Population} generated flexible augmentation policy schedules to speed up the searching procedure (5000 GPU hours to 5 GPU hours on CIFAR-10). Peng et al. \cite{Peng2018Jointly} augmented samples by adversarial learning with pre-training processes.

With respect to text recognition, the training of the recognizer requires much data. The widely used synthetic datasets \cite{gupta2016synthetic,Jaderberg2015Reading} provide more than 10 million samples. However, Li et al. \cite{li2018show} additionally used approximately 50k public real datasets for training and significantly improved recognition performance, which suggests that the recognition models are still data-hungry. As for handwritten text, existing training data can hardly cover various writing styles and generating synthetic handwritten data is also challenging. Unlike scene text synthesis, there is few font in writing style to render on a canvas.

Our method is proposed for multiple characters augmentation in an automatic manner. An agent network searches hard training samples online. Moreover, the framework is end-to-end trainable without any fine-tuning. 

\section{Methodology}
\subsection{Overall Framework}
As illustrated in Figure \ref{pic-overall}, the proposed framework consists of three main modules: an agent network, an augmentation module and a recognition network. First, we initialize a set of custom fiducial points on the image. A moving state predicted by the agent network and a randomly generated moving state are fed to the augmentation module. The moving state indicates the movement of a set of custom fiducial points. Then the augmentation module takes the image as input, and applies transformation based on the moving states respectively. The recognizer predicts text strings on the augmented images. Finally, we measure the recognition difficulty of the augmented images under the metric of edit distance. The agent learns from the moving state that increases difficulty, and explores the weakness of the recognizer. As a result, the recognizer gains robustness from the hard training samples.

As we only use the prediction of the recognition network and the difficulty is measured by edit distance rather than other loss functions, the recognition network can be replaced by recent advanced methods \cite{shi2017end,shi2018aster}, which we will demonstrate in the section \ref{section-exp}. In this section, we describe the augmentation module and the joint training scheme of the proposed framework.

\subsection{Text Augmentation} \label{section-text-augment} 
\begin{figure}[t]
\centering
\includegraphics[width=0.8\columnwidth]{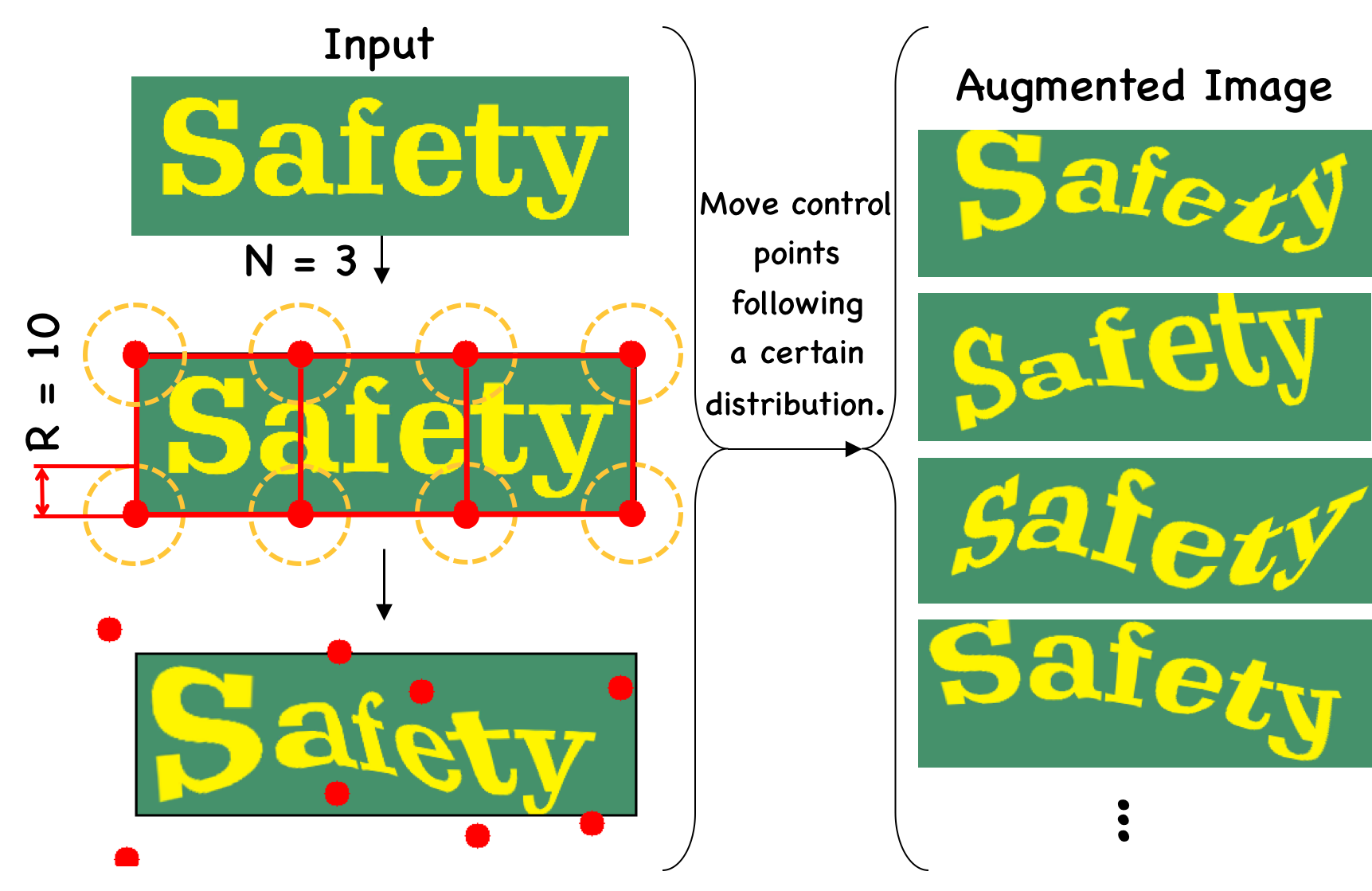} 
\caption{Text augmentation. The image is divided into three patches ($N=3$) and the moving radius is limited to ten ($R=10$). The red points denote control points. }
\label{pic-similar-trans}
\end{figure}

Given a text image, the augmentation goal is to increase the diversity of every character in the text string. This motivates us to use more custom fiducial points for transformation. As shown in Figure \ref{pic-similar-trans}, we averagely divide the image into $N$ patches and initialize $2(N+1)$ fiducial points $p$ along the top and bottom image borders. After that, we augment images by following a certain distribution and randomly moving the fiducial points to $q$ within the radius $R$. 

To generate an augmented image, we apply similarity deformation based on moving least squares \cite{Schaefer2006Image} on the input image. Given a point $u$ in the image, the transformation for $u$ is

\begin{equation}
T(u)=\left(u-p_{*}\right)\textbf{M}+q_{*},
\end{equation}
where $\textbf{M} \in \mathbb{R}^{2\times2}$ is a linear transformation matrix that is constrained to have the property $M^\mathrm{T}M=\lambda^2I$ for some scalar $\lambda$. Here $p_{*}$ and $q_{*}$ are the weighted centroids of initialized fiducial points $p$ and moved fiducial points $q$, respectively:
 
\begin{equation}
p_{*}=\frac{\sum_{i=1}^{2(N+1)} w_{i} p_{i}}{\sum_{i=1}^{2(N+1)} w_{i}},  q_{*}=\frac{\sum_{i=1}^{2(N+1)} w_{i} q_{i}}{\sum_{i=1}^{2(N+1)} w_{i}}.
\end{equation}

The weight $w_{i}$ for point $u$ has the form
\begin{equation}
w_{i}=\frac{1}{\left|p_{i}-u\right|^{2 \alpha}}, u \ne p_{i}.
\end{equation}

Note that as $u$ approaches $p_{i}$, the weight $w_{i}$ increases. This means that $u$ mostly depends on the movement of the nearest  fiducial point. The $w_{i}$ is bounded. If $u = p_{i}$, then $T(u) = u$. Here we set $\alpha=1$.

The best transformation $T(u)$ is obtained by minimizing
\begin{equation}
\sum \nolimits_{i=1}^{2(N+1)} w_{i}\left|T_{u}\left(p_{i}\right)-q_{i}\right|^{2},
\end{equation}
to yield the unique minimizer \cite{Schaefer2006Image}. 

\begin{figure}[b]
\centering
\includegraphics[width=0.9\columnwidth]{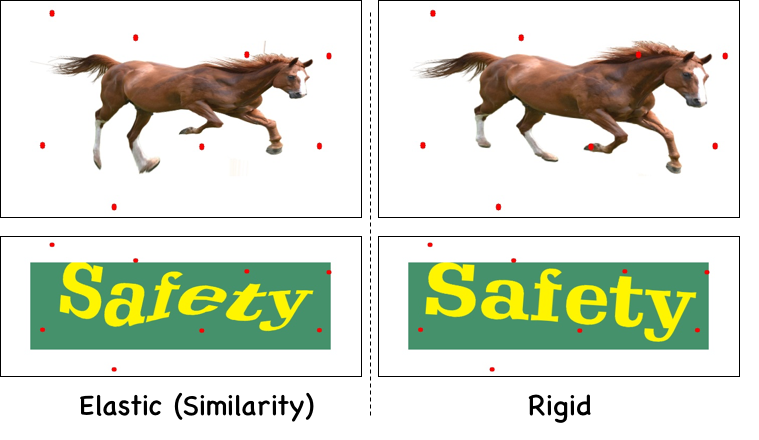} 
\caption{Comparison of the elastic (similarity) and rigid transformation. The movements of the fiducial points on all the images are the same. The rigid transformation retains relative shape (realistic for general object), but text image augmentation requires more flexible deformation for every character. Therefore, the elastic (similarity) transformation is more suitable for text image augmentation.}
\label{pic-rigid}
\end{figure}

\textbf{Discussion} Though Thin Plate Spline Transformation (TPS) \cite{bookstein1989principal} has achieved success in shape rectification \cite{shi2018aster} and feature-level adversarial learning \cite{bhunia2019handwriting}, it is reported that TPS appears non-uniform scaling and shearing, which is undesirable in many applications \cite{Schaefer2006Image}. 
One possible reason why previous work used TPS may be all the operators in TPS are differentiable and can be found in most mainstream deep learning libraries. 
As the learning of our augmentation is free of backward calculation of recognition loss, and our goal is to setup a general augmentation, we choose similarity deformation based on moving least squares as our transformation strategy. Besides, we also compare similarity transformation with rigid transformation \cite{Schaefer2006Image}, which is regarded as the most realistic transformation for general object. As illustrated in Figure \ref{pic-rigid}, the rigid transformation retains relative shape (realistic for general object), but the similarity transformation is more suitable for text image augmentation, because it provides more flexible deformation for every character. Further analysis is given in Section \ref{section-ablation-study} and Table \ref{table-ablation-method}.

\subsection{Learnable Agent} 

Different from the previous smart augmentation method \cite{Cubuk2019AutoAugment} that used reinforcement learning to search for best policies, we solve the learning problem in a faster and more efficient fashion. Inspired by heuristic algorithms, we find solutions among all possible ones. As the training procedure is dynamic, approximate solutions are sufficient and exact solutions are computationally expensive. For every step in the training procedure, we generate a variation of the predicted moving state. It serves as a candidate of learning target. If the random moving state increases recognition difficulty, then the agent learns from the moving state. In contrast, we reverse the learning target if the moving state decreases recognition difficulty. 

We formulate the problem of finding harder distorted sample as a movement learning problem. As illustrated in Figure \ref{pic-similar-trans}, given an image, we randomly move the fiducial points to warp the image. The moving operation $(\Delta x, \Delta y)$ for every fiducial point is associated with two factors: 1) the direction of movement, namely, the signs of $(\Delta x, \Delta y)$; 2) the distance of movement, namely, $(|\Delta x|, |\Delta y|)$. In our practice, the learning of distance fails to converge. It is hard for the agent network to precisely learn the distance of the movement. Another interesting observation is that the failed agent network always predicts maximum moving distance to create excessive distorted samples, which reduced the stability of recognizer training. Therefore, we limit the learning space to the direction of movement. Based on the moving direction, the moving distance is randomly generated within the range of radius. It avoids tedious movement predicted by the agent network, because the randomness introduces uncertainties in the augmentation. Moreover, the agent network can be designed as a lightweight architecture. As shown in Table \ref{table-agent}, the agent network consists of only six convolutional layers and a fully connected layer. The storage requirement of the agent network is less than 1.5M.

The learning scheme of the agent network is shown in Algorithm \ref{alg-agent}. First, the learnable agent predicts a moving state distribution aiming to create a harder training sample. A random moving state is also fed to the augmentation module. Then the augmentation module generates augmented samples based on the two moving state, respectively. After that, the recognition network takes the augmented samples as input and predicts text strings. The difficulty of the pair of samples is measured by the edit distance between the ground truth and predicted text strings. Finally, the agent takes the moving state that increase difficulty as guidance and updates itself. The unified framework is end-to-end trainable.

\begin{algorithm}[t]
\caption{Joint Learning Scheme}
\label{alg-agent}
\hspace*{0.02in} {Input image} $I_{in}$ and {Ground truth} $GT$;\\
\hspace*{0.02in} {Patch number} $N$ and {Moving radius} $R$;\\
\hspace*{0.02in} {Initialized fiducial points} $p$.
\begin{algorithmic}[1]
\State Sample moving state as $S$ from predicted distribution:
\begin{center} $S=\textbf{Agent}(I_{in})$.\end{center}
\State Generate random moving state $S'$ (randomly select one point in $S$ and switch to the opposite direction).
\State Both $S$ and $S'$ contain direction for movement.
\State Within the range of $R$, randomly move $p$ based on $S$ and $S'$ to obtain $q$ and $q'$, respectively.
\begin{center} $I_{Aug}=\textbf{Augment}(I_{in}, p, q)$, \end{center}
\begin{center} $I_{Aug}'=\textbf{Augment}(I_{in}, p, q')$. \end{center}
\State Recognize $I_{Aug}$ and $I_{Aug}'$:
\begin{center} $Reg=\textbf{Recognizer}(I_{Aug})$,\end{center}
\begin{center} $Reg'=\textbf{Recognizer}(I_{Aug}')$.\end{center}
\State Update \textbf{Recognizer} using $I_{Aug}$.
\State Measure difficulty by edit distance $\textbf{ED}(\cdot)$:
\If{$\textbf{ED}(Reg, GT) \le \textbf{ED}(Reg', GT)$}
	\begin{center} $S'$ increases recognition difficulty. \end{center}
	\begin{center} Update \textbf{Agent} network with $S'$ by minimizing: \end{center}
	\begin{equation}
	Loss=-\sum_{i=1}^{2(N+1)} \log \big(P(S_{i}' | I_{in})\big)
	\end{equation}
\Else 
	\begin{center} Update \textbf{Agent} network with reversed direction $-S'$ \\ by minimizing: \end{center}
	\begin{equation}
	Loss=-\sum_{i=1}^{2(N+1)} \log \big(P(-S_{i}' | I_{in})\big)
	\end{equation}
\EndIf
\end{algorithmic}
\end{algorithm}

\begin{table}[b]
\renewcommand\arraystretch{1.1}
\centering
\caption{Architecture of the agent network. ``AP" denotes $2\times2$ average pooling. ``BN" represents batch normalization. The kernel size, stride and padding size of all the convolutional layers are 3, 1 and 1, respectively. The output size means 2(N + 1) points, two coordinates and two moving directions.}
\label{table-agent}
\begin{small}
\begin{tabular}{c|c}
\hline
Type  & Size \\
\hline
Input & $1\times32\times100$ \\
\hline
Conv-16, ReLU, AP & $16\times16\times50$ \\
\hline
Conv-64, ReLU, AP & $64\times8\times25$ \\
\hline
Conv-128, BN, ReLU & $128\times8\times25$ \\
\hline
Conv-128, ReLU, AP & $128\times4\times12$ \\
\hline
Conv-64, BN, ReLU & $64\times4\times12$ \\
\hline
Conv-16, BN, ReLU, AP & $16\times2\times6$ \\
\hline
FC-8(N+1) & $8(N+1)$ \\
\hline
Reshape & $2(N+1)\times2\times2$ \\
\hline
\end{tabular}
\end{small}
\end{table}

\section{Experiments}
\label{section-exp}
In this section, we conduct extensive experiments on various benchmarks, including regular and irregular scene text, and handwritten text. We first conduct ablation studies to analyze the impact of the size of training data, the number of divided patches $N$ and the moving radius $R$ on performance. Our method is also compared to existing affine and rigid transformations. Then we integrate state-of-the-art recognition models with our method to show the effectiveness of our learnable data augmentation. Finally, we combine our method with the feature-level adversarial learning method \cite{bhunia2019handwriting} to further boost the recognition performance, which suggests that our method is flexible and can be applied in other augmentation systems.

\subsection{Scene Text Datasets}

The widely used synthetic datasets \cite{Jaderberg2015Reading} and \cite{gupta2016synthetic} contain 9-million and 8-million synthetic words respectively. We randomly sample 10k, 100k and 1 million images (refered to as \textbf{Syn-10k}, \textbf{Syn-100k} and \textbf{Syn-1m respectively}) for ablation studies.

\textbf{Real-50k} is collected by Li et al. \cite{li2018show} from all the public real datasets, containing approximately 50k samples. 

IIIT 5K-Words \cite{mishra2012scene} (\textbf{IIIT5K}) contains 3000 cropped word images for testing. 

Street View Text \cite{wang2011end} (\textbf{SVT}) consists of 647 word images for testing. Many images are severely corrupted by noise and blur. 

ICDAR 2003 \cite{lucas2003icdar} (\textbf{IC03}) contains 867 cropped images after discarding images that contained non-alphanumeric characters or had fewer than three characters \cite{wang2011end}. 

ICDAR 2013 \cite{karatzas2013icdar} (\textbf{IC13}) inherits most of its samples from IC03. It contains 1015 cropped images.

Street View Text Perspective \cite{quy2013recognizing} (\textbf{SVT-P}) contains 645 cropped images for testing. Most of them are perspective distorted. 

CUTE80 \cite{risnumawan2014robust} (\textbf{CT80}) contains 80 high-resolution images taken in natural scenes. It was specifically collected to evaluate the performance of curved text recognition. It contains 288 cropped natural images.

ICDAR 2015 \cite{karatzas2015icdar} (\textbf{IC15}) is obtained by cropping the words using the ground truth word bounding boxes and includes more than 200 irregular text images. 

\subsection{Handwritten Text Datasets}

\textbf{IAM} \cite{marti2002iam} contains more than 13,000 lines and 115,000 words written by 657 different writers.

\textbf{RIMES} \cite{augustin2006rimes} contains more than 60,000 words written in French by over 1000 authors. 

\subsection{Implementation Details}

\textbf{Network} The architecture of the agent network is detailed in Table \ref{table-agent}, which is a lightweight network (less than 1.5M) consisting of six convolutional layers and a fully connected layer. The output size means $2(N+1)$ points, two coordinates and two moving directions. As we use the edit distance as the metric of difficulty, the framework is independent of various recognition losses. For instance, Shi et al. \cite{shi2017end} adopted CTC loss \cite{graves2006connectionist} for convolutional recurrent neural network and the attentional decoders \cite{cluo2019moran,shi2018aster} are guided by the cross-entropy loss. Therefore, our framework is friendly to different recognizers. We show the flexibility of our method in the following experiments. 

\textbf{Optimization} 
In the ablation study, we use ADADELTA \cite{zeiler2012adadelta1} with default learning rate as the optimizer. The batch size is set to 64. All the images are resized to $(32, 100)$. When our method is integrated with recent state-of-the-art recognizers, the experiment settings, including optimizer, learning rate, image size, and training and testing datasets, are the same as those of the recognizers for the sake of fair comparison. 

\textbf{Environment} All experiments are conducted on NVIDIA 1080Ti GPUs. The augmentation module takes less than 2ms to generate a $(32, 100)$ image on a 2.0GHz CPU. It is possible to take advantage of multi-threaded acceleration. For every iteration, the end-to-end training with learnable augmentation takes less than 1.5 times of the training time of the single recognizer. If it is trained with random augmentation, there is nearly no extra time consumption.

\subsection{Ablation Study}
\label{section-ablation-study}
In this section, we perform a series of ablation studies. As the released scene text datasets \cite{gupta2016synthetic,Jaderberg2015Reading} provide tens of millions of training samples, it is possible to sample small datasets with three orders of scales. Therefore, we conduct ablation studies on scene text datasets. The training datasets are Real-50k, Syn-10k, Syn-100k and Syn-1m. We use ADADELTA \cite{zeiler2012adadelta1} with default learning rate as the optimizer. The batch size is set to 64. All the images are resize to $(32, 100)$. In Table \ref{table-ablation-method}, we combine all the scene text testing sets as a unified large dataset for evaluation. 

As the attentional recognizer is the most cutting-edge method, we choose the network equipped with ResNet and attentional decoder in \cite{shi2018aster} as the recognizer. The recognizer trained without any augmentation serves as a baseline. Following the widely used evaluation metric \cite{cluo2019moran,shi2018aster}, the performance is measured by word accuracy in Table \ref{table-ablation-method}-\ref{table-ablation-radius}. To ensure that the training is sufficient, we train the models 10 more epochs after they achieve highest accuracy.

\textbf{Size of Training data}
As shown in Table \ref{table-ablation-method}, the recognizer using our learnable augmentation method outperforms the baseline by a large margin. For instance, the largest margin of 14.0\% is on the Syn-10k dataset. This suggests that our proposed method greatly improves the generalization of recognizer in small-data settings. With the increase of the dataset size, the gap reduces. But there is still a significant accuracy increase of 6.5\% on the one million training data Syn-1m. 

\begin{table}[t]
\renewcommand\arraystretch{1.1}
\centering
\caption{Ablation studies on the size of training data and transformation with the settings of $N = 3$ and $R = 10$. ``Aug." denotes our augmentation method under a randomly initialized distribution for direction sampling. }
\label{table-ablation-method}
\begin{small}
\setlength{\tabcolsep}{1.9mm}{
\begin{tabular}{c|c|c|c|c}
\hline
Method  & Real-50k & Syn-10k & Syn-100k & Syn-1m \\
\hline
\hline
baseline & 54.1 & 7.7 & 39.5 & 60.9 \\
\hline
Affine & 58.6 & 16.9 & 43.9 & 61.7 \\
\hline
Rigid & 58.7 & 17.5 & 44.9 & 63.9 \\
\hline
\hline
Aug. & 63.4 & 20.1 & 48.6 & 65.9 \\
\hline
Aug.+Agent & \textbf{66.5} & \textbf{21.7} & \textbf{51.2} & \textbf{67.4} \\
\hline
\end{tabular}
}
\end{small}
\end{table}

\begin{figure}[b]
\centering
\includegraphics[width=8cm, height=4cm]{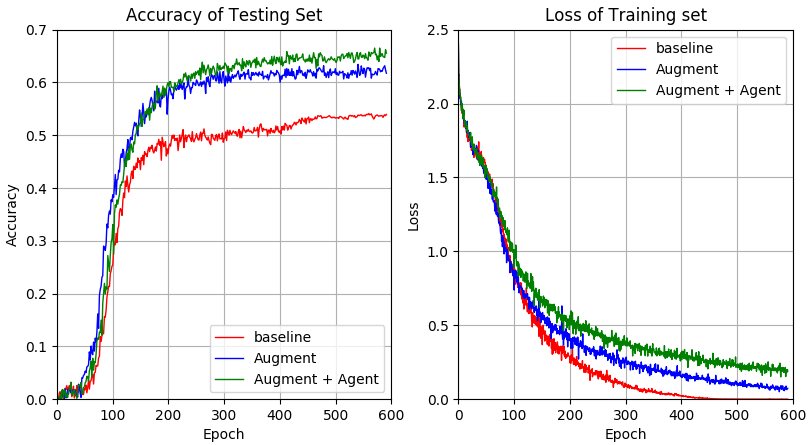} 
\caption{Training loss on Real-50k and testing accuracy on the large evaluation dataset.}
\label{pic-acc-loss}
\end{figure}

\textbf{Transformation}
Affine transformation \cite{jaderberg2015spatial} including rotation, scaling and translation, is compared with our augmentation method in Table \ref{table-ablation-method}. The results show that the recognizer using affine augmentation outperforms the baseline but still falls behind the recognizer that uses our augmentation method, because the affine transformation is limited to designed geometric deformations, which are unable to cover the diversity of text appearance. We also conduct an experiment to study the effectiveness of the rigid transformation. As discussed in Section \ref{section-text-augment}, although the rigid transformation is realistic for general object \cite{Schaefer2006Image}, the similarity transformation is more suitable for text image augmentation.

\textbf{Learnable Agent}
In Table \ref{table-ablation-method}, the agent network further boosts the performance by jointly learning data augmentation and recognizer training. In particular, it achieves an accuracy increase of 3.1\% when the recognizer is trained using Real-50k. The curves of training loss on Real-50k and testing accuracy on the large evaluation dataset are illustrated in Figure \ref{pic-acc-loss}. An interesting observation is that the loss of the recognizer with learnable agent decreases slower than others, which suggests that the agent network explores the weakness of the recognizer and generates harder samples for training. Thus the recognizer keeps learning and gains robustness. In contrast, the traditional recognizer stops learning when the loss is close to zero. 

\textbf{Patch Number and Moving Radius}
We study two key parameters $N$ and $R$ respectively. The training dataset is Syn-10k. Table \ref{table-ablation-patch} and Table \ref{table-ablation-radius} show the experiment results. We find that for regular text, to achieve the best performance, the patch number $N$ can be set to $2$ or $3$. As for irregular text (SVT-P, CT80 and IC15), it is better to set $N$ to $3$, because under this setting, numerous curve text images are generated for training. The recognizer thus gains robustness. We further illustrate the effectiveness of the variance of moving radius $R$ in Table \ref{table-ablation-radius}. The best setting for a $(32, 100)$ image is $R=10$. In the following experiments, we use the best setting for $N$ and $R$ for further studies.

\begin{table}[t]
\renewcommand\arraystretch{1.1}
\centering
\caption{Ablation studies on the number of patches. $R$ is set to 10.}
\label{table-ablation-patch}
\begin{small}
\setlength{\tabcolsep}{1.8mm}{
\begin{tabular}{c|c|c|c|c|c|c|c}
\hline
$N$  & IIIT5K & SVT & IC03 & IC13 & SVT-P & CT80 & IC15 \\
\hline
\hline
1 & 23.5 & 6.6 & 19.6 & 22.3 & 6.0 & 10.4 & 10.6 \\
\hline
2 & \textbf{29.8} & 10.5 & \textbf{29.3} & 29.3 & 8.2 & 14.6 & \textbf{14.3} \\
\hline
3 & 29.4 & \textbf{10.8} & 27.2 & \textbf{29.6} & \textbf{9.1} & \textbf{16.3} & \textbf{14.3} \\
\hline
4 & 26.5 & 7.3 & 22.6 & 25.6 & 5.8 & 11.5 & 11.0 \\
\hline
5 & 26.1 & 7.4 & 22.6 & 26.9 & 6.0 & 13.5 & 11.2 \\
\hline
\end{tabular}
}
\end{small}
\end{table}

\begin{table}[t]
\renewcommand\arraystretch{1.1}
\centering
\caption{Ablation studies on the moving radius. $N$ is set to 3.}
\label{table-ablation-radius}
\begin{small}
\setlength{\tabcolsep}{1.8mm}{
\begin{tabular}{c|c|c|c|c|c|c|c}
\hline
$R$  & IIIT5K & SVT & IC03 & IC13 & SVT-P & CT80 & IC15  \\
\hline
\hline
0 & 10.9 & 2.3 & 9.0 & 13.0 & 1.8 & 5.2 & 3.6 \\
\hline
2 & 13.4 & 2.2 & 9.8 & 14.2 & 2.0 & 5.2 & 4.3 \\
\hline
5 & 20.3 & 4.6 & 17.0 & 20.4 & 4.2 & 9.0 & 7.8 \\
\hline
10 & \textbf{29.4} & \textbf{10.8} & \textbf{27.2} & \textbf{29.6} & \textbf{9.1} & \textbf{16.3} & \textbf{14.3} \\
\hline
15 & 28.8 & 8.3 & 26.1 & 27.8 & 6.3 & 13.2 & 12.2 \\
\hline
\end{tabular}
}
\end{small}
\end{table}

\subsection{Integration with State-of-the-art Methods}

\begin{figure}[t]
\centering
\includegraphics[width=0.9\columnwidth]{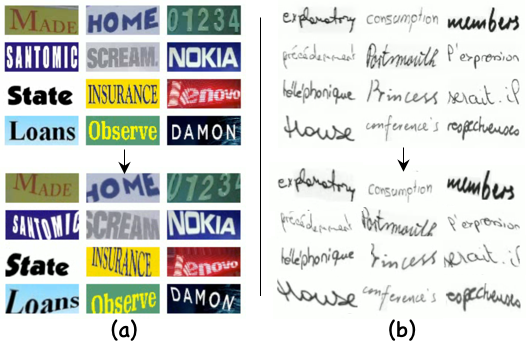} 
\caption{Visualization of augmented samples on (a) scene text and (b) handwritten text.}
\label{pic-demo}
\end{figure}

In this section, we integrate our proposed method with state-of-the-art recognizers. The augmented samples for different tasks are shown in Figure \ref{pic-demo}. We first show the improvement of attention-based recognizer \cite{shi2018aster} on irregular scene text benchmarks. Then we validate the generalization of our method by using CTC-based recognizer \cite{bhunia2019handwriting} and conducting experiments on handwritten text. Note that our method automatically adapt to general text recognition tasks without any manual modification. Moreover, we show that our method is flexible and can be integrated with other augmentation systems to further boost the performance.

\begin{table}[b]
\renewcommand\arraystretch{1.}
\centering
\caption{Word accuracy on irregular text. ``*" denotes the result is from one rectification iteration for fair comparison.}
\label{table:Results on general benchmarks}
\begin{tabular}{c|c|c|c }
\toprule
\multirow{2}{*}{Method} & \multicolumn{3}{c}{Irregular Text} \\
\cline{2-4}
& SVT-P & CT80 & IC15 \\
\midrule
Shi, Bai, and Yao \cite{shi2017end} & 66.8 & 54.9 & -\\
Shi et al. \cite{shi2016robust} & 71.8 & 59.2 & -\\
Liu et al. \cite{liu2016star} & 73.5 & - & - \\
Yang et al. \cite{yang2017learning} & 75.8 & 69.3 & -\\
Cheng et al. \cite{cheng2017focusing} & 71.5 & 63.9 & 70.6 \\
Liu, Chen, and Wong \cite{liu2018char} & - & - & 60.0\\
Cheng et al. \cite{cheng2017arbitrarily} & 73.0 & 76.8 & 68.2 \\
Bai et al. \cite{bai2018edit} & - & - & 73.9\\
Liu et al. \cite{liu2018synthetically} & 73.9 & 62.5 & - \\
Luo, Jin, and Sun \cite{cluo2019moran} & 76.1 & 77.4 & 68.8 \\
Liao et al. \cite{liao2019scene} & - & 78.1 & - \\
Shi et al. \cite{shi2018aster} & 78.5 & 79.5 & \textbf{76.1}\\
Zhan and Lu \cite{Zhan2018ESIR}* & 77.3 & 78.8 & 75.8\\
\midrule
baseline (ASTER \cite{shi2018aster}) & 77.7 & 79.9 & 75.8 \\
+ Ours & \textbf{79.2} & \textbf{84.4} & \textbf{76.1} \\
\bottomrule
\end{tabular}
\end{table}

\textbf{Irregular Scene Text Recognition}
Irregular shape is one of the challenges for scene text recognition. ASTER proposed by Shi et al. \cite{shi2018aster} is an attention-based recognizer equipped with rectification network. We study the robustness of the recognizer by augmenting training samples and increasing the diversity of text appearance. The experiment settings, including optimizer, learning rate, image size, and training datasets, are the same as ASTER \cite{shi2018aster}.

The performance improved by our method is compared to state-of-the-art methods. Although using real samples \cite{li2018show} and character-level geometric constraints \cite{yang2019symmetry} to train the recognizer can significantly improve the performance, we follow the setting of most methods for fair comparison. As Zhan and Lu \cite{Zhan2018ESIR} rectified images for several times and Shi et al. \cite{shi2018aster} only performed rectification once, we choose the result with one rectification iteration reported in the paper. The performance of scene text recognizers is measured by word accuracy.

As shown in Table \ref{table:Results on general benchmarks}, we first reproduce the same recognizer as ASTER \cite{shi2018aster}, which serves as a baseline. The results of the reimplemented ASTER are comparable to the results in the original paper. Then we integrate our method with the recognizer. A significant accuracy gain occurs on CT80 (4.5\%). It is noteworthy that there is still a notable improvement (1.5\%) on SVT-P, which contains images with noise, blur and low-resolution. Though abundant synthetic samples may cover a lot of variation of text appearance, our augmentation shows reasonable improvement on irregular text recognition. The result is competitive with recent state-of-the-art methods. 

\begin{table}[b]
  \centering
  \caption{Comparison with previous methods on IAM. AFDM is the key module of \cite{bhunia2019handwriting}. }
    \begin{tabular}{c|c|c|c|c}
    \toprule
    \multirow{2}[2]{*}{Method} & \multicolumn{2}{c|}{Unconstrained} & \multicolumn{2}{c}{Lexicon} \\
    \cline{2-5}
          & WER   & CER   & WER   & CER \\
    \midrule
    Bosquera et al. \cite{espana2010improving} & -     & -     & 20.01 & 11.27 \\
    Almaz{\'a}n et al. \cite{almazan2014word} & -     & -     & 15.50 & 6.90 \\
    Sun et al. \cite{sun2016convolutional}  & -     & -     & 11.51 & - \\
    Sueiras et al. \cite{sueiras2018offline} & 23.80 & 8.80  & 19.70 & 9.50 \\
    Ptucha et al. \cite{ptucha2019intelligent} & - & -  & 8.22  & 4.70 \\
    Zhang et al. \cite{zhang2019sequence} & 22.20 & 8.50  & -     & - \\
    Bhunia et al. \cite{bhunia2019handwriting} & 17.19 & 8.41  & 8.87  & 5.94 \\
    \midrule
    baseline & 19.12 & 7.39  & 10.07 & 5.41 \\
    + Ours & 14.04 & 5.34  & 7.52  & 3.82 \\
    + AFDM \cite{bhunia2019handwriting} & 16.40 & 6.40  & 8.77  & 4.67 \\
    + Ours + AFDM \cite{bhunia2019handwriting} & \textbf{13.35} & \textbf{5.13} & \textbf{7.29} & \textbf{3.75} \\
    \bottomrule
    \end{tabular}%
  \label{IAM_Result}%
\end{table}%

\textbf{Handwritten Text Recognition} 
As the diversity of handwriting styles is the main challenge of handwritten text recognition \cite{almazan2014word} and limited training data is difficult to cover all handwriting styles, we evaluate our model on two popular datasets IAM \cite{marti2002iam} and RIMES \cite{augustin2006rimes} to validate the effectiveness of our method. We use Character Error Rate (CER) and Word Error Rate (WER) as metrics for handwritten text recognition. The CER measures the Levenshtein distance normalized by the length of the ground truth. The WER denotes the ratio of the mistakes at the word level, among all words of the ground truth.

We compare our method to state-of-the-art methods in the Table \ref{IAM_Result} and Table \ref{RIMES_Result}. Besides, a comparison with previous augmentation method of Bhunia et al. \cite{bhunia2019handwriting} is conducted. For fair comparisons, our experiment settings are the same with \cite{bhunia2019handwriting}.

We apply the same CTC-based recognition network as \cite{bhunia2019handwriting}. The baseline shown in Table \ref{IAM_Result} and Table \ref{RIMES_Result} is the reproduced result. Further, we reproduce Adversarial Feature Deformation Module (AFDM) \cite{bhunia2019handwriting} in the recognition network. The AFDM is the key module proposed by Bhunia et al. \cite{bhunia2019handwriting} for smart augmentation. The accuracy increases as expected. Note that our reproduced results are better than most of the results (7 of 8) in the original paper, which verifies the effectiveness of our implementations and experiments. We find that our augmentation greatly contributes to the robustness of the recognizer. It improves the performance by a large margin (5.08\% unconstrained WER reduction on IAM) and significantly performs better than AFDM. The recognizer trained using our method also outperforms all the state-of-the-art methods.

\begin{table}[t]
  \centering
  \caption{Comparison with previous methods on RIMES. AFDM is the key module of \cite{bhunia2019handwriting}. }
    \begin{tabular}{c|c|c|c|c}
    \toprule
    \multirow{2}[2]{*}{Method} & \multicolumn{2}{c|}{Unconstrained} & \multicolumn{2}{c}{Lexicon} \\
    \cline{2-5}
          & WER   & CER   & WER   & CER \\
    \midrule
    Sueiras et al. \cite{sueiras2018offline} & 15.90 & 4.80  & 13.10 & 5.70 \\
    Ptucha et al. \cite{ptucha2019intelligent} & - & -  & 5.68  & 2.46 \\
    Bhunia et al. \cite{bhunia2019handwriting} & 10.47 & 6.44  & 6.31  & 3.17 \\
    \midrule
    baseline & 13.83 & 3.93  & 4.94  & 2.02 \\
    + Ours & 9.23  & 2.57  & 4.41  & 1.49 \\
    + AFDM \cite{bhunia2019handwriting} & 11.81 & 3.33  & 4.85  & 1.92 \\
    + Ours + AFDM \cite{bhunia2019handwriting} & \textbf{8.67} & \textbf{2.42} & \textbf{3.90} & \textbf{1.37} \\
    \bottomrule
    \end{tabular}%
  \label{RIMES_Result}%
\end{table}%

Finally, we use both AFDM and our method for training and further boost the performance of the recognizer by a notable accuracy increase. This suggests that our method is a meta framework, which can be applied in other augmentation systems. 

\section{Conclusion}

In this paper, we propose a learnable augmentation method for the training of text recognizer. Our method may be the first geometric augmentation method specifically designed for sequence-like characters. Furthermore, our method bridges the gap between the data augmentation and network optimization by joint learning. The proposed method is simple yet effective. It is able to automatically adapt to general text recognition tasks without any manual modification. Extensive experiments show that our method boosts the performance of the recognizers for both scene text and handwritten text. Moreover, our method is a meta framework that potentially can be incorporated into other augmentation systems. In future, we will extend our method for more general applications in multiple object detection and recognition.

\section*{Acknowledgement}
This research is supported in part by NSFC (Grant No.: 61936003), the National Key Research and Development Program  of China (No. 2016YFB1001405), and GD-NSF (no.2017A030312006).

{\small
\bibliographystyle{ieee_fullname}
\bibliography{mybibfile.bib}
}

\end{document}